%% file: ms.tex
\documentclass[graybox]{svmult}

\usepackage{mathptmx}       
\usepackage{helvet}         
\usepackage{courier}        
\usepackage{type1cm}        
\usepackage{makeidx}         
\usepackage{graphicx}        
\usepackage{multicol}        
\usepackage{multirow}
\usepackage[bottom]{footmisc}

\usepackage{amsmath, amssymb}

\makeindex             

\begin{document}

\title*{Multiplicative Latent Force Models}
\author{Daniel J. Tait and Bruce J. Worton}
\institute{Daniel J. Tait \at School of Mathematics, University of Edinburgh \email{Tait.djk@gmail.com}
\and Bruce J. Worton \at School of Mathematics, University of Edinburgh \email{Bruce.Worton@ed.ac.uk}}
%
%
\maketitle

\abstract*{Bayesian modelling of dynamic systems must achieve a compromise between providing a complete mechanistic specification of the process while retaining the flexibility to handle those situations in which data is sparse relative to model complexity, or a full specification is hard to motivate. Latent force models achieve this dual aim by specifying an evolution equation consisting of a simple linear evolution equation combined with an additive latent Gaussian process (GP) forcing term.\newline\indent
  In this paper we extend the latent force framework to allow for multiplicative interactions between the GP and the latent states leading to more control over the geometry of the trajectories. Unfortunately inference is no longer straightforward and so we introduce an approximate density based on the method of successive approximations and examine the performance of our method by way of a simulation study.}

\abstract{Bayesian modelling of dynamic systems must achieve a compromise between providing a complete mechanistic specification of the process while retaining the flexibility to handle those situations in which data is sparse relative to model complexity, or a full specification is hard to motivate. Latent force models achieve this dual aim by specifying a parsimonious linear evolution equation which an additive latent Gaussian process (GP) forcing term.\newline\indent
  In this work we extend the latent force framework to allow for multiplicative interactions between the GP and the latent states leading to more control over the geometry of the trajectories. Unfortunately inference is no longer straightforward and so we introduce an approximation based on the method of successive approximations and examine its performance using a simulation study.}

\section{Introduction}
\label{sec:intro}
Modern statistical inference must often achieve a balance between an appeal to the \emph{data driven paradigm} whereby models are sufficiently flexible so as to allow inference to be chiefly driven by the observations, and on the other hand the \emph{mechanistic approach} whereby the structure of the data generating process is well specified up to some, usually modest, set of random parameters. The conflict between these two philosophies can be particularly pronounced for complex dynamic systems for which a complete mechanistic description is often hard to motivate and instead we would like a framework that allows for the specification of a, potentially over-simplistic, representative evolution equation which would enable the modeller to embed as much prior knowledge as they feel comfortable doing while at the same time ensuring the model is sufficiently flexible to allow for any unspecified dynamics to be captured during the inference process.

Such a compromise is provided by a class of hybrid models introduced in \cite{alvarez} which they term \emph{latent force models} (LFM). This is a combination of a simple mechanistic model with added flexibility originating from a flexible Gaussian process forcing term. The aim is to encode minimal dynamic systems properties into the resulting state trajectories without necessarily having to provide a complete mechanistic description of how the system evolves.

One of the appealing features of the latent force model is the fact that the resulting trajectories are given by Gaussian processes and therefore inference can proceed in a straightforward manner. Nevertheless for many classes of systems in which we might be interested this assumption of Gaussian trajectories is unlikely to be realistic; examples of this might include time series of wind direction data which are necessarily distributed on the unit circle, motion data with restricted degrees of freedom and tensor valued data in geophysical systems. For all of these cases if we have a suitably dense sample then the Gaussian trajectory assumption may be acceptable, however when data are sparse comparative to model complexity we would like to be able to consider models that move beyond this Gaussian trajectory assumption and allow a priori embedding of geometric constraints.

In this paper we briefly review the latent force model before introducing our extension in Section \ref{sec:mlfm} and then discuss how our model now allows for the embedding of strong geometric constraints. Unfortunately it is no longer straightforward to solve for the trajectories as some transformation of the latent random variables and therefore in Section \ref{sec:succapprox} we introduce an approximate solution method for this class of models based on the method of successive approximations for the solution of certain integral equations. Finally, in Section \ref{sec:sim} we demonstrate by way of simulation that our approximate model performs well for cases which possess a solvable ground truth model.

\section{Latent Force Models}
The latent force model initially inspired by the problem of modelling transcriptional regulation of gene activities \cite{gao, lawrence} before the underlying model philosophy shifted from this mechanistic perspective to the hybrid setting in \cite{alvarez}. For a $K$-dimensional state variable $\mathbf{x}(t) \in \mathbb{R}^K$ the first order latent force model is described by a system of ordinary differential equations in matrix-vector form as
\begin{align}
  \frac{\operatorname{d}\mathbf{x}(t)}{\operatorname{d}t} = - \mathbf{D}\mathbf{x}(t) + \mathbf{b} + \mathbf{S}\mathbf{g}(t), \label{eq:lfm}
\end{align}
where $\mathbf{D}$ is a $K \times K$ real-valued diagonal matrix, $\mathbf{b}$ is a real-valued $K$-vector and $\mathbf{g}(t)$ is the $\mathbb{R}^R$-valued stochastic process with smooth independent Gaussian process components $g_r(t)$, $r=1,\ldots, R$. The $K \times R$ rectangular sensitivity matrix $\mathbf{S}$ acts to distribute linear combinations of the independent latent forces to each component of the evolution equation. 

The model \eqref{eq:lfm} gives only an implicit link between the latent random variables and the observed trajectories, but to carry out inference we would ideally like to represent this connection as an explicit transformation. It turns out that for the model \eqref{eq:lfm} with constant coefficient matrix and additive inhomogeneous forcing term this is easily done and an explicit solution is given by
\begin{align}
  \mathbf{x}(t) = e^{-\mathbf{D}(t-t_0)}\mathbf{x}(t_0) + \int_{t_0}^te^{-\mathbf{D}(t-\tau)}\operatorname{d}\tau \cdot \mathbf{b} + L[\mathbf{g}](t), \label{eq:lfm_sol}
\end{align}
where $L[f](t)$ is the linear integral transformation acting on functions $f  : \mathbb{R} \rightarrow \mathbb{R}^R$ to produce a function $L[f] : \mathbb{R} \rightarrow \mathbb{R}^K$ given by
\begin{align}
  L[f](t) = \int_{t_0}^t e^{-\mathbf{D}(t-\tau)}\mathbf{S} f(\tau) \operatorname{d}\tau. \label{eq:lfm_integral_op}
\end{align}
The decomposition of the solution of the latent force model \eqref{eq:lfm_sol} makes it clear that, for given values of the initial condition $\mathbf{x}(t_0)$ and the model parameters $\boldsymbol{\theta} = (\mathbf{D}, \mathbf{b}, \mathbf{S})$, the trajectory is given by a linear integral transformation of the latent Gaussian processes, and it follows that the trajectory and the latent force variables will have a joint Gaussian distribution. From this we may deduce that the LFM is a particular form of Gaussian process regression model with hyperparameters, $\boldsymbol{\theta}$, along with any additional hyperparameters of the latent Gaussian force terms, and as such inference for the latent variables and any hyperparameters may be done using standard Gaussian process regression methods \cite{gpml}.
\section{Multiplicative Latent Force Models}
\label{sec:mlfm}
While from a computational point of view the Gaussian process regression framework of the LFM is appealing we would like to move beyond the restrictive assumption of having Gaussian trajectories for the state variables. We therefore introduce an extension of the LFM which will allow us to represent non-Gaussian trajectories while at the same time keeping the same fundamental components of the latent force model: a linear ODE with the time dependent behaviour of the evolution equation coming from a set of independent smooth latent Gaussian process forces. In matrix/vector form our model is given by
\begin{align}
  \frac{\operatorname{d}\mathbf{x}(t)}{\operatorname{d}t} = \mathbf{A}(t)\mathbf{x}(t), \qquad \mathbf{A}(t) = \mathbf{A}_0 + \sum_{r=1}^R \mathbf{A}_r \cdot g_r(t). \label{eq:mlfm}
\end{align}
The coefficient matrix $\mathbf{A}(t)$ will be a square matrix of dimension $K \times K$ formed by taking linear combinations of a set of \emph{structure matrices} $\{\mathbf{A}_r\}_{r=0}^R$ which we multiply by scalar Gaussian processes. By linearity $\mathbf{A}(t)$ will be a Gaussian process in $\mathbb{R}^{K \times K}$ although typically the choice of the set of structure matrices will be guided by geometric considerations and in general the dimension of this space will be much less than that of the ambient $K^2$ dimensional space.

In the specification \eqref{eq:mlfm} the matrix valued Gaussian process $\mathbf{A}(t)$  will interact multiplicatively with the state variable in the evolution equation, rather than as an additive forcing term in \eqref{eq:lfm}, and so we refer to this model as the \emph{multiplicative latent force model} (MLFM). The existence of solutions to ODEs of the form \eqref{eq:mlfm} where the coefficient matrix is a stochastic process are discussed in \cite{strand}.

The multiplicative interaction in \eqref{eq:mlfm} and the freedom to choose the support of the coefficient matrix will allow us to embed strong geometric constraints on solutions to ODEs of this form. In particular by choosing the elements $\{A_r\}$ from some Lie algebra $\mathfrak{g}$ corresponding to a Lie group $G$ then the fundamental solution of \eqref{eq:mlfm} will itself be a member of the group $G$ \cite{iserles}, allowing the creation of dynamic models with trajectories either within the group itself or formed by an action of this group on some vector space.
\section{Method of Successive Approximations}
\label{sec:succapprox}
In general non-autonomous linear ODEs do not possess a closed form solution and therefore it is no longer straightforward to carry out inference for the MLFM; we lack the explicit representation of the trajectories in terms of the latent random processes which was possible for the LFM using the solution \eqref{eq:lfm_sol}. To proceed we first note that a pathwise solution to the model \eqref{eq:mlfm} on the interval $[0, T]$ is given by the integral equation
\begin{align*}
  \mathbf{x}(t) = \mathbf{x}(0) + \int_{0}^t \mathbf{A}(\tau)\mathbf{x}(\tau)\operatorname{d}\tau, \qquad 0 \leq t \leq T,
\end{align*}
a solution to which can be obtained by starting from an initial approximation of the trajectory, $\mathbf{x}_0(t)$, and then repeatedly iterating the linear integral operator
\begin{align}
  \mathbf{x}_{m+1}(t) = \mathbf{x}_0(0) + \int_{0}^{t} \mathbf{A}(\tau)\mathbf{x}_m(\tau)\operatorname{d}\tau. \label{eq:picard}
\end{align}
This process is known as the \emph{method of successive approximations} and is a classical result in the existence and uniqueness theorems for the solutions of ODEs.

We introduce some probabilistic content in to this approximation by placing a mean zero Gaussian process prior on the initial state variable $\mathbf{x}_0(t)$ independent of the latent force terms. Since \eqref{eq:picard} is a linear operator for known $\mathbf{A}(t)$ and $\mathbf{x}_m(t)$ then the marginal distribution of the $(m+1)$th successive approximation conditional on the process $\mathbf{A}(t)$ will be mean zero Gaussian with covariance given recursively by
\begin{align}
  \tilde{\mathbb{E}}\left[\mathbf{x}_{m+1}(t)\mathbf{x}_{m+1}(t')^{\top}\right] = \int_{t_0}^t \int_{t_0}^{t'} \mathbf{A}(\tau) \mathbb{E}[\mathbf{x}_{m}(\tau)\mathbf{x}_m(\tau')^{\top}]\mathbf{A}(\tau')^{\top}\operatorname{d}\tau\operatorname{d}\tau', \label{eq:sa_marginal_moments}
\end{align}
where $\tilde{\mathbb{E}}$ denotes expectation conditional on the stochastic process $\mathbf{A}(t)$ on $[0, T]$.
  
In practice we will not be dealing with complete trajectories, but instead with the process observed at a finite set of points $t_0 < \cdots < t_N$, and so we replace the map \eqref{eq:picard} by a numerical quadrature
\begin{align}
  \mathbf{x}(t_0) + \int_{t_0}^{t_i} \mathbf{A}(\tau)\mathbf{x}(\tau) \D\tau \approx \mathbf{x}(t_0) + \sum_{j=1}^{N_i} \mathbf{A}(\tau_{ij})\mathbf{x}(\tau_{ij})w_{ij} \label{eq:discrete_picard}, \qquad i=1,\ldots, N,
\end{align}
for a set of weights $\{w_{ij}\}$ which are determined by our choice of quadrature rule and we have a set of nodes $\tau_{ij}$ labelled such that $\tau_{i1} = t_{i-1}$ and $\tau_{iN_i} = t_i$. It follows that methods with more than two nodes over a particular interval $[t_i, t_{i+1}]$ must necessarily augment the latent state vector. Increasing the number of nodes will cause the error in \eqref{eq:discrete_picard} to decrease, we defer discussion of the finer points of this approximation, but for practical purposes the important detail is that this error can be made arbitrarily small because we are free to increase the resolution of the trajectories by treating this as a missing data problem albeit with a corresponding computational cost. In terms of a linear operator acting on the whole trajectory we replace the operator \eqref{eq:picard} with a matrix operator $K[\mathbf{g}]$ acting on the discrete trajectories such that each row of $K[\mathbf{g}]$ performs the quadrature \eqref{eq:discrete_picard}, that is if $\mathbf{x}$ is a dense realisation of a continuous process $\mathbf{x}(t)$ evaluated at the points $\{ \tau_{ij} \}$ then
\begin{align}
  \left( \mathbf{K}[\mathbf{g}]\mathbf{x} \right)_i = \mathbf{x}(t_0) + \sum_{j=1}^{N_i}\mathbf{A}(\tau_{ij})\mathbf{x}(\tau_{ij})w_{ij}, \qquad i=1,\ldots,N. \label{eq:row_discrete_picard}
\end{align}
For suitably dense realisations of the trajectory we can conclude that the majority of the informational content in the linear map \eqref{eq:picard} is captured by applying the matrix operator form of the integral operator \eqref{eq:row_discrete_picard} and therefore there will be minimal loss of information if we replace the (Gaussian) correlated error term with an \emph{independent} additive noise term leading to a conditional distribution of the form
\begin{align}
  p(\mathbf{x}_{m+1} \mid \mathbf{x}_m, \mathbf{g}, \boldsymbol{\Gamma} ) &= \mathcal{N}\left(\mathbf{x}_{m+1} \mid K[\mathbf{g}]\mathbf{x}_m, \boldsymbol{\Gamma}\right), \label{eq:approx_cond_dist}
\end{align}
for some covariance matrix $\boldsymbol{\Gamma}$ parameterising the additive noise error. A similar use of quadrature is proposed in \cite{lawr:eff_gpsampling} applied to the integral operator \eqref{eq:lfm_integral_op} to allow for nonlinear transformation of the GP variables, no attempt is made to proxy for the quadrature error and it effectively gets absorbed into the GP model, for our application the additive error may be viewed as a regularisation term to prevent singularities in the covariance matrix. Heuristically in the limit with $\Gamma = 0$ and $M \rightarrow \infty$ the covariance matrix can be represented as the outer product of the $K$ eigenvectors of the discretised matrix operator $K[\mathbf{g}]$ with unit eigenvalues so that the resulting covariance matrix is singular.

If we specify a Gaussian initial distribution $p(\mathbf{x}_0) = \mathcal{N}(\mathbf{x}_0 \mid \mathbf{0}, \Sigma_0)$ then carry out iterates of the map \eqref{eq:discrete_picard} up to some truncation order $M$ we have an approximation to the distribution of a finite sample of a complete trajectory of \eqref{eq:mlfm} conditioned on a discrete realisation of the latent forces which is given by
\begin{align}
  p(\mathbf{x}_M \mid \mathbf{g}, \boldsymbol{\Gamma}) &= \int \cdots \int p(\mathbf{x}_M, \mathbf{x}_{M-1}, \ldots, \mathbf{x}_0 \mid \mathbf{g}, \boldsymbol{\Gamma} ) \D \mathbf{x}_0 \cdots \D \mathbf{x}_{M-1}\notag \\
  &= \int \cdots \int \prod_{m=1}^{M} p(\mathbf{x}_m \mid \mathbf{x}_{m-1}, \mathbf{g}, \boldsymbol{\Gamma} ) p(\mathbf{x}_0) \D\mathbf{x}_0 \cdots \D \mathbf{x}_{M-1} \notag \\    
  &= \mathcal{N}(\mathbf{x}_M \mid \mathbf{0}, \Sigma_M(\mathbf{g}, \boldsymbol{\Gamma}) ), \label{eq:marginal_Mth_order_sa}
\end{align}
where the covariance matrix $\Sigma_M(\mathbf{g}, \boldsymbol{\Gamma})$ is defined recursively by $\Sigma_0(\mathbf{g}, \boldsymbol{\Gamma}) = \Sigma_0$ and 
\begin{align}
  \Sigma_{m}(\mathbf{g}, \boldsymbol{\Gamma}) = K[\mathbf{g}]\Sigma_{m-1}(\mathbf{g}, \boldsymbol{\Gamma})K[\mathbf{g}]^{\top} + \Gamma, \qquad m=1, \ldots, M, \label{eq:sa_cov_mat}
\end{align}
and this model should then be viewed as a discretisation of the true marginal distribution with moments \eqref{eq:sa_marginal_moments}.

It is now possible to specify a complete joint distribution $p(\mathbf{x}, \mathbf{g})$ of the latent state and force variables by completing the likelihood term \eqref{eq:marginal_Mth_order_sa} with the prior on the latent force variable. On inspection of \eqref{eq:row_discrete_picard} we see that the entires of $K[\mathbf{g}]$ will be linear in the latent force variables and therefore the entries of the covariance matrix \eqref{eq:sa_cov_mat} will be degree $2M$ polynomials in the latent force variables and as such there is no analytical expression for the posterior conditional density for orders greater than one. Despite this it is straightforward to use to use sampling methods and gradient based approximations.

\section{Simulation Study}
\label{sec:sim}
Reasonably we would expect that by increasing the truncation order of the approximation introduced in the previous section we gain increasingly accurate approximations to the true conditional distribution and in this section we demonstrate that this is indeed the case by considering an exactly solvable model.

We demonstrate our method on the Kubo oscillator \cite{risken} which can be expressed by the ODE in $\mathbb{R}^2$ with a single latent force and evolution equation
\begin{align}
  \begin{bmatrix}
    \dot{x}(t) \\
    \dot{y}(t)
  \end{bmatrix} &=
  \begin{bmatrix}
    0 & - g(t) \\ g(t) & 0 
  \end{bmatrix}
  \begin{bmatrix}
    x(t) \\ y(t)
  \end{bmatrix}, \label{eq:kubo}
\end{align}
which for $\mathbf{x}(t) = (x(t), y(t))^{\top}$ has solution given by
\begin{align}
  \mathbf{x}(t) = R\left[\int_{0}^{t}g(\tau)\D\tau\right]\mathbf{x}(t_0), \label{eq:kubo_sol}
\end{align}
where $R[\theta]$ in \eqref{eq:kubo_sol} is the $2 \times 2$ matrix rotating a vector in $\mathbb{R}^2$ by $\theta$-radians anticlockwise around the origin. It follows that given a set of data points $\mathcal{Y} = (\mathbf{x}_0,\mathbf{x}_1,\ldots,\mathbf{x}_N)$ with $t_0 < t_1 < \cdots < t_N$ and zero measurement error that the values of $G_i := \int_{t_{i-1}}^{t_i} g(\tau)\D \tau$ are constrained to satisfy $\mathbf{x}_i = R[G_i]\mathbf{x}_{i-1}$, for $i=1,\ldots, N$ which defines the vector $\mathbf{G} = (G_1, \ldots, G_N)^{\top}$ up to translation of each component by $2\pi$, moreover since $\operatorname{Var}(G_i) = \mathcal{O}(|t_i-t_{i-1}|^2)$ we can consider only the component in $[-\pi, \pi]$ and approximate the true conditional distribution of $\mathbf{g} = (g(t_0), g(t_1), \ldots, g(t_N))^{\top}$ by the Gaussian distribution with density $p(\mathbf{g} \mid \mathbf{G} = \boldsymbol{\gamma})$ with
\begin{align*}
  \boldsymbol{\gamma} = \left\{ \boldsymbol{\gamma} \in [-\pi, \pi]^N \; : \; \mathbf{x}_i = R[\gamma_i]\mathbf{x}_{i-1}, \; i=1,\ldots, N \right\}.
\end{align*}

While the distribution implied by the likelihood term \eqref{eq:marginal_Mth_order_sa} is not available in closed form, nevertheless we can investigate the qualitative properties of the method introduced in Section \ref{sec:succapprox} by considering the Laplace approximation. Using the Laplace approximation has the benefit of allowing us to carry out the comparison with the ground truth distribution using a proper metric on the space of distributions by considering the Wasserstein distance between two multivariate Gaussians \cite{dowson}.

The method of successive approximations fixes a point and is therefore necessarily local in character, as such we implement a simulation study that enables us to assess the performance of our approximation as the total interval length increases. We consider two methods of varying the interval length $T$; the first by fixing the sample size, $N$, and then varying the spacing between samples, $\Delta t$, and the second by fixing the sample frequency and varying the total number of observations. For each combination of sample size and frequency we perform $100$ simulations of the Kubo oscillator \eqref{eq:kubo} on the interval $[0, T]$ assuming a known \emph{radial basis function} (RBF) kernel $k(t, t' ; \psi) = \psi_0\exp\{(t-t')^2 / 2\psi_1^2\}$ with $\psi = (1, 1)^{\top}$ for the latent force. We consider interval lengths $T \in \{3, 6, 9\}$ and sample frequencies $\Delta t \in \{0.50, 0.75, 1.00\}$. This implies a sample size of $N = T / \Delta t + 1$ for each experiment and we use a Simpson's rule to perform the quadrature \eqref{eq:discrete_picard} so that the latent state vector is augmented to the size $2N + 1$.

Our principal interest is in the impact of the truncation order, $M$, on the accuracy of our approximation and so for each simulated experiment we fit the model with orders $M = 3, 5, 7, 10$. The covariance matrix $\boldsymbol{\Sigma}_0$ of the initial approximation is formed by placing indendent Gaussian process priors on the first and second components with RBF kernels $k(t, t' ; \phi_k)$ and the parameters $\phi_k$, $k=1,2$ are optimised during the fitting process. The regularisation matrix $\boldsymbol{\Gamma}$ is giving by multiplying an appropriately size identity matrix by a small scale parameter $0.0001$ and this value is kept fixed during the optimisation process.

The results of the experiment are displayed in Table \ref{tab:simresults}, along each row we observe that across all sampling specifications that increasing the order of approximation leads to increasingly accurate approximations of the true distributions, and that this conclusion holds whether we vary the sample size or the sample frequency. Inspecting the columns we observe that for each order a decrease in the sampling interval $T$ leads to a a general increase in accuracy of the approximation with some variations with the sample size and frequency. The fact that within most blocks of fixed $T$ and $M$ that the distances are of a similar magnitude strongly suggests that it is the size of the window $T$ that is a larger determinant of the accuracy of the introduced approximation than the number of sample points or their frequency. In fact we see that dense samples can lead to a slower convergence of the approximation and this is particularly pronounced for the row $T=9.$ and $\Delta t = 0.50$ which does a very poor job of approximating the true distribution at lower orders compared to the sparser samples, but eventually outperforms these methods as the approximation order increases.

\begin{table}
  \caption{Comparison of the successive approximations MLFM model introduced in the text with the true distribution for the Kubo oscillator based on $100$ simulations of the process on $[0, T]$ with $N = T / \Delta t + 1$ evenly spaced observations. Reported are the sample averages and standard errors of the Wasserstein distance between the Laplace approximation and the true conditional distribution}
  \label{tab:simresults}
  \begin{tabular}{p{1.0cm}p{1.5cm}p{2.2cm}p{2.2cm}p{2.2cm}p{2.2cm}}  
  \hline\noalign{\smallskip}
  $T$ & $\Delta t$ & order=3 & order=5 & order=7 & order=10  \\  
  \noalign{\smallskip}\svhline\noalign{\smallskip}
  $9$ & $1.00$ & $0.965$ $(0.477)$ & $0.863$ $(0.573)$ & $0.711$ $(0.672)$ & $0.527$ $(0.632)$ \\
  & $0.75$ & $0.983$ $(0.315)$ & $0.874$ $(0.407)$ & $0.762$ $(0.448)$ & $0.584$ $(0.415)$ \\
  & $0.50$ & $1.517$ $(0.556)$ & $1.068$ $(0.450)$ & $0.701$ $(0.227)$ & $0.517$ $(0.225)$ \\
    $6$ & $1.00$ & $0.865$ $(0.606)$ & $0.619$ $(0.503)$ & $0.433$ $(0.475)$ & $0.319$ $(0.412)$ \\
  & $0.75$ & $0.738$ $(0.392)$ & $0.629$ $(0.463)$ & $0.513$ $(0.426)$ & $0.328$ $(0.325)$ \\
  & $0.50$ & $0.846$ $(0.256)$ & $0.591$ $(0.194)$ & $0.532$ $(0.234)$ & $0.399$ $(0.192)$ \\
    $3$ & $1.00$ & $0.374$ $(0.311)$ & $0.294$ $(0.384)$ & $0.202$ $(0.256)$ & $0.185$ $(0.211)$ \\
  & $0.75$ & $0.421$ $(0.440)$ & $0.272$ $(0.440)$ & $0.136$ $(0.217)$ & $0.076$ $(0.064)$ \\
  & $0.50$ & $0.421$ $(0.190)$ & $0.395$ $(0.289)$ & $0.235$ $(0.132)$ & $0.191$ $(0.051)$ \\  

  \noalign{\smallskip}\hline\noalign{\smallskip}
\end{tabular}
\end{table}

\section{Discussion}
In this paper we have introduced the MLFM, a hybrid model which enables the embedding of prior geometric knowledge into statistical models of dynamic systems. By using the method of successive approximations we were able to motivate a family of truncated approximations to the joint distribution, and while the distribution is not available in closed form it is still amenable to sampling and gradient based methods. In future work we discuss variational sampling methods formed by retaining the full set of successive approximations rather than performing the marginalisation \eqref{eq:marginal_Mth_order_sa} and exploiting the interpretation of \eqref{eq:approx_cond_dist} as a linear Gaussian dynamical system in the truncation order.

The simulation study in Section \ref{sec:sim} showed the method performs well over moderate sample windows with only a few orders of approximation, but that as the length of window over which the solution was being solved increases the order required to achieve good performance increases. It may therefore be of interest to replace a single, high order, approximation with a collection of lower order local approximations. Combining these local models in a principled manner is the subject of ongoing work, nevertheless the results of Section \ref{sec:sim} show that the method introduced in this paper show this method can perform well, as well as being an important precursor to more involved methods.

\begin{acknowledgement}
Daniel Tait is supported by an EPSRC studentship.
\end{acknowledgement}
\newpage
\input{refs}

\end{document}

%% file: refs.tex
%
%
%